\documentclass[journal]{IEEEtran}
\usepackage{xcolor,soul,framed} 
\colorlet{shadecolor}{yellow}
\usepackage[pdftex]{graphicx}
\graphicspath{{../pdf/}{../jpeg/}}
\DeclareGraphicsExtensions{.pdf,.jpeg,.png}
\usepackage[cmex10]{amsmath}
\usepackage{array}
\usepackage{mdwmath}
\usepackage{mdwtab}
\usepackage{eqparbox}
\usepackage{url}
\usepackage{fancyhdr}
\usepackage{hyperref}
\usepackage{amssymb}

\usepackage{booktabs}
\usepackage{multirow}
\usepackage{graphicx}
\usepackage[ruled,linesnumbered]{algorithm2e}

\begin{document}

\markboth{\parbox[t]{20cm}{This work has been submitted to the IEEE for possible publication.\newline Copyright may be transferred without notice, after which this version may no longer be accessible.}}{}

\bstctlcite{IEEEexample:BSTcontrol}
    \title{Rule-driven News Captioning}
  \author{
  Ning~Xu,
  Tingting~Zhang,
  Hongshuo~Tian, 
  An-An~Liu$^\ast$
  
  \thanks{N.~Xu, T.~Zhang, H.~Tian and A.-A.~Liu are with the School of Electrical and Information Engineering, Tianjin University, Tianjin 300072, China.
(Corresponding author: A.-A.~Liu, E-mail: anan0422@gmail.com).
}
  \thanks{Manuscript received XXX, 202X; revised XXX, 202X. Copyright \copyright 20xx IEEE. Personal use of this material is permitted. However, permission to use this material for any other purposes must be obtained from the IEEE by sending an email to pubs-permissions@ieee.org.}
}

\maketitle

\begin{abstract}
News captioning task aims to generate sentences by describing named entities or concrete events for an image with its news article. 
Existing methods have achieved remarkable results by relying on the large-scale pre-trained models, which primarily focus on the correlations between the input news content and the output predictions. 
However, the news captioning requires adhering to some fundamental rules of news reporting, such as accurately describing the individuals and actions associated with the event. 
In this paper, we propose the rule-driven news captioning method, which can generate image descriptions following designated rule signal. 
Specifically, we first design the news-aware semantic rule for the descriptions. 
This rule incorporates the primary action depicted in the image (e.g., ``performing") and the roles played by named entities involved in the action (e.g., ``Agent" and ``Place").  
Second, we inject this semantic rule into the large-scale pre-trained model, BART, with the prefix-tuning strategy, where multiple encoder layers are embedded with news-aware semantic rule.
Finally, we can effectively guide BART to generate news sentences that comply with the designated rule. 
Extensive experiments on two widely used datasets (i.e., GoodNews and NYTimes800k) demonstrate the effectiveness of our method.

\end{abstract}

% === KEYWORDS ====================================================================
% =================================================================================

\IEEEpeerreviewmaketitle

% ====================================================================
% ====================================================================
% ====================================================================

\section{Introduction}

Traditional image captioning task \cite{WuZSZCGSJ22}, \cite{XianLTM22} aims to produce general captions, which overlook named entities in images, such as specific people, organizations, and places. 
In contrast, the news captioning task \cite{TranMX20}, \cite{ZhangFMZ022} leverages background knowledge from relevant news articles to generate named-entity-included captions, thereby greatly improving the quality and practicality of image captions.

\begin{figure}[!t]
\centering
\includegraphics[width=\linewidth]{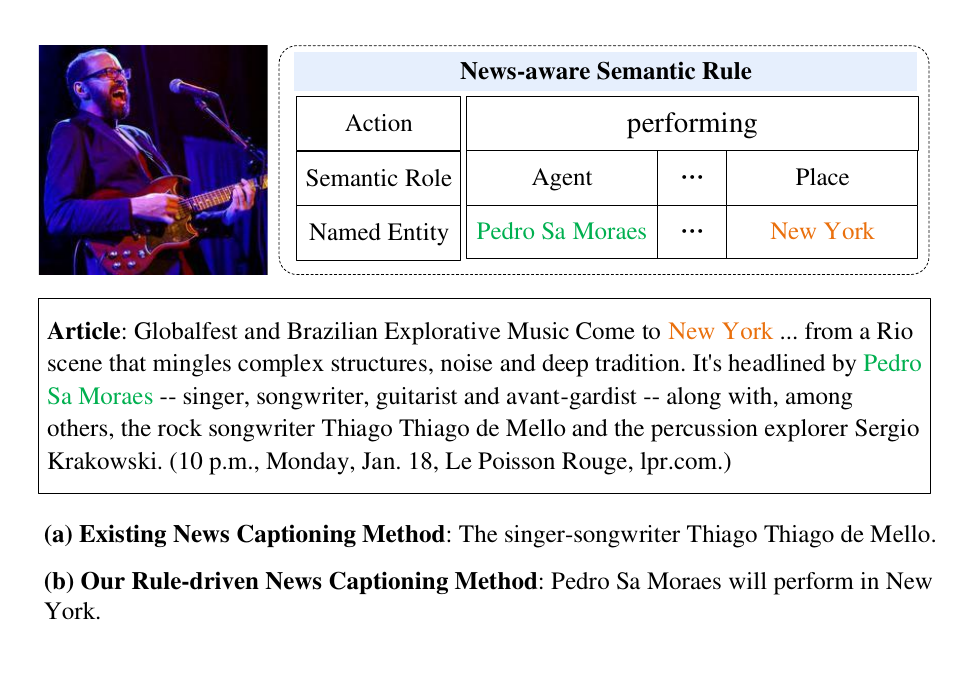}
\caption{Comparison between existing news captioning method and our method. 
We design the news-aware semantic rule that includes the primary action depicted in the image as well as the roles played by the named entities involved in the action, which guides the model to generate the news sentence that adheres to the rules of news reporting. }
\label{example_diagram}
\end{figure}

Significant progress has been made in news captioning community. 
For examples, Biten et al. \cite{BitenGRK19} generate a template with placeholders, and then select named entities from news articles to replace the placeholders.
Tran et al. \cite{TranMX20} use the byte-pair encoding to generate unseen or rare named entities.
Recently, there have been several attempts to apply large-scale pre-trained models to the task of news captioning. 
For examples, Zhang et al. \cite{ZhangFMZ022} utilize a multi-modal entity prompting mechanism to fine-tune the large-scale pre-trained model. 
However, these methods still face a significant challenge. 
The news captioning task requires adhering to some fundamental rules of news reporting, such as accurately describing the individuals, actions, or locations for the special event. 
The existing methods ignore the rich rule patterns, which only focus on the correlation between input news content and output predictions. 
As shown in Fig. \ref{example_diagram}, given an image with a man who is performing and the news article, the existing captioner may generate the rule-agnostic sentence ``The singer-songwriter Thiago Thiago de Mello", which can not strictly describe the location ``New York" and the correct person's name ``Pedro Sa Moraes".

To tackle this problem, we propose the rule-driven news captioning method, which can generate image descriptions following designated rule signal. Our method consists of three components: 
\textbf{1) Named Entity Extraction.} 
We use the pre-trained named entity recognition (NER) technique to extract named entities from news articles. 
Then, the CLIP model is employed to compute the image-text similarity to match the named entities that closely align with the news image content. To ensure that the CLIP model is applicable to the news domain, we fine-tuned it on the news caption dataset.
\textbf{2) News Rule Construction.} 
We design the news-aware semantic rule for the descriptions. 
This rule incorporates the primary action depicted in the image (e.g., ``performing”) and the set of roles played by named entities surrounding the action (e.g., ``Agent” and ``Place”). Meanwhile, it can present a clear affiliation between generic objects (e.g., ``man") in images and named entities (e.g., ``Pedro Sa Moraes") in news articles.
\textbf{3) Caption Generation.} 
We incorporate the news-aware semantic rule into the large-scale pre-trained model, BART, using the prefix-tuning strategy. 
Specifically, BART can be used as the pre-trained  multi-modal summarization model \cite{LewisLGGMLSZ20} to generate captions for both images and news articles. 
We embed the semantic rule into BART's multiple encoder layers. As a result, BART is effectively guided to generate news sentences that adhere to the rules of news reporting. 

We conduct comprehensive experiments on two publicly available news datasets, i.e., GoodNews \cite{BitenGRK19} and NYTimes800k \cite{TranMX20}. 
Experimental results show that our method can achieve competitive or state-of-the-art performance against the existing methods on all benchmarks. 
Especially under CIDEr metric, our method can achieve significant improvement over the best competitor \cite{ZhangFMZ022} by 7.59\% on GoodNews. 
Ablation studies further examine the efficacy of each component. 
Besides, we visualize the generated news-aware semantic rules, to clearly observe the named entities' changes improved by our method. 
In summary, we have the following main contributions: 

\begin{enumerate}
\item We propose the novel rule-driven news captioning method. Different from existing works that primarily focus on the relationship between input news content and output predictions, our approach enables the captioner to follow designated rule signal for the caption generation. 
\item We design the news-aware semantic rule that encompasses the primary action depicted in the image as well as the roles played by the named entities involved in the action. This semantic rule is then injected into the large-scale pre-trained model, BART, using the prefix-tuning strategy, which guides BART to generate news sentences. 
\item Extensive experiments are conducted on two widely used datasets, including GoodNews \cite{BitenGRK19} and NYTimes800k \cite{TranMX20}, to assess the effectiveness of our method. The visualization of the semantic rules confirms its capacity to control the named entities within news descriptions. 
\end{enumerate}

\section{Related Work}

\subsection {General Image Captioning}
The image captioning task is a challenging task that combines computer vision and natural language processing technology \cite{CaoAZW22, VinyalsTBE15, LiZLY19, YanHLYLMCG22}. 
Vinyals et al. \cite{VinyalsTBE15} first use CNN to encode the image, and then use LSTM to generate the corresponding image description.
Yao et al. \cite{YaoPLM18} propose GCN \cite{zhu2022interpretable}, \cite{gan2022multigraph} plus LSTM.
Li et al. \cite{LiZLY19} adopt a framework based on CNN-Transformer. They introduce EnTangled Attention in Transformer to integrate visual and semantic information more effectively.
Compared with the CNN-Transformer framework, Liu et al. \cite{abs-2101-10804} achieve more efficient global context modeling through a full Transformer model.
% Wang et al. \cite{9834140} propose CrowdCaption, a crowd scenes caption dataset, and devise a method for CrowdCaption that can generate crowd-specific detailed descriptions.
Zhang et al. \cite{ZhangSLJZWHJ21} introduce the Adaptive-Attention module in the Transformer decoder to measure the contribution of visual cues and language cues to caption generation.
Wu et al. \cite{WuZSZCGSJ22} use segmentation features as additional information sources to enhance the contribution of visual information in the prediction process.
Cornia et al. \cite{abs-2111-12727} design Universal Captioner, which adopts network-level noise data for training and uses the long-tail concept to describe images.
% Jing et al. \cite{10183355} utilize the implicit external knowledge contained in the dataset to generate video captions.
Although traditional image captioning task performs very well, this method can only generate general descriptions and lacks specific contextual information \cite{YuLYH20, YanHLYLMCG22, XuMYTWJ22}.
In comparison, our method uses image-article pairs as input, and with the help of the background knowledge contained in the news articles, it can generate descriptions containing named entities, which enhances the practicality and accuracy of the descriptions.

\subsection {News Image Captioning}
In recent years, news image captioning task has received increasing attention \cite{abs-2108-02050}, \cite{TranMX20}, \cite{ZhouLRY22}. Compared with the traditional image captioning task, the news image captioning task can effectively combine the image content and the background knowledge of the news article to generate captions containing named entities. 
Early research work often generates captions in two steps \cite{BitenGRK19}, \cite{RamisaYMM18}.
Firstly, a caption with placeholders is generated, then the correct named entities are selected from news articles to replace the placeholders in the caption.
For example, Biten et al. \cite{BitenGRK19} use different types of named entities as placeholders to generate the template caption, e.g., PER performs in GPE on DATE. Then they select named entities from the sentences with the highest attention values in the news articles to fill these placeholders.

Recently, an increasing number of works have adopted end-to-end models. 
% Hu et al. \cite{abs-2108-02050} implement a text information aggregation approach from the sentence-level to the word-level to generate entity-aware captions. 
Tran et al. \cite{TranMX20} utilize RoBERTa \cite{abs-1907-11692} and ResNet-152 \cite{HeZRS16} to encode news articles and images respectively, and handle uncommon words by using byte-pair encoding in the Transformer decoder.
Zhao et al. \cite{abs-2107-11970} propose to mine external web knowledge to construct a multi-modal knowledge graph, thereby associating visual objects with named entities.
Yang et al. \cite{YangKTJ21} design a set of template components based on the six elements of news (i.e., who, when, where, what, why, how) to guide the model in generating news captions that comply with journalistic guidelines. 
Zhang et al. \cite{ZhangFMZ022} utilize a multi-modal entity prompting mechanism to fine-tune the large-scale pre-trained model. 
% Zhou et al. \cite{ZhouLRY22} generate accurate captions by combining key local context containing important named entities with global context that can describe the entire article.

Although existing methods have achieved significant results, they overlook the fundamental rules of news reporting, focusing only on the correlation between the input news content and the output prediction.
In contrast, we propose a rule-driven news captioning method that can generate descriptions based on designated rule signal.

\begin{figure*}[!t]
  \begin{center}
  \includegraphics[width=\linewidth]{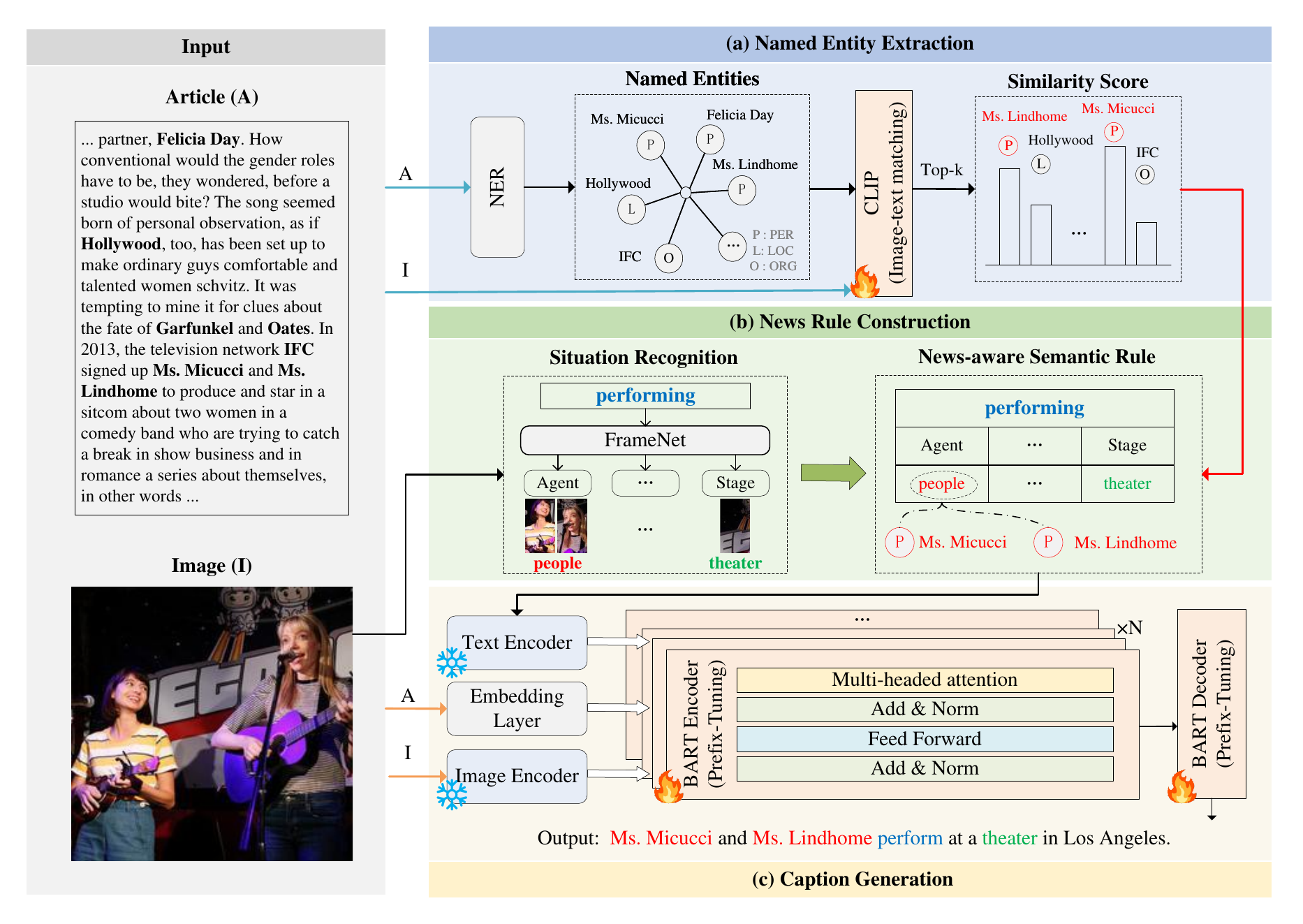}\\
  \caption{Overview of the proposed method.  
  \textbf{(a) Named Entity Extraction.} 
 We use the Named Entity Recognition (NER) model to extract named entities and their categories from news articles. Then, the fine-tuned CLIP model is applied to calculate the image-text similarity score, filtering the named entities accordingly. 
 \textbf{(b) News Rule Construction.} 
We design the news-aware semantic rule, which includes the primary action depicted in the image
 (e.g., ``performing”) and the named entities involved in the action (e.g., ``Ms. Micucci”, ``Ms. Lindhome” and ``theater”) along with their corresponding roles (e.g., ``Agent” and ``Stage”). 
 \textbf{(c) Caption Generation. }
 We integrate the news-aware semantic rule, news article, and image, into BART, a large-scale pre-trained model, using the prefix-tuning strategy, for the news caption generation. }\label{model_diagram}
  \end{center}
\end{figure*}

\subsection {Large-Scale Pre-trained Models}
Large-scale pre-trained models have received widespread attention \cite{abs-2005-14165}, \cite{abs-2302-13971}, \cite{abs-2204-02311}, \cite{abs-2302-05442}, \cite{abs-2304-02643}, \cite{RadfordKHRGASAM21}, \cite{LewisLGGMLSZ20}.
These models are usually pre-trained on large-scale datasets, and then fine-tuned for specific tasks to better adapt to various complex application scenarios. 
They have demonstrated excellent results in tasks such as common sense reasoning \cite{abs-2005-14165}, \cite{abs-2302-13971} and visual question answering \cite{abs-2303-01903}, \cite{YangGW0L0W22}.
Among them, CLIP \cite{RadfordKHRGASAM21} leverages 400 million image-text pairs gathered from the internet, and uses contrastive learning \cite{peng2023grlc}, \cite{mo2023multiplex} to achieve cross-modal semantic alignment between image features and text features.
BART \cite{LewisLGGMLSZ20} is a large-scale pre-trained language model based on Transformer.
Our approach enables news description generation by injecting news-aware semantic rule into multiple encoder layers of BART, using the prefix-tuning strategy.

\subsection {Situation Recognition}
Situation recognition aims to describe the main activities depicted in an image by using a semantic framework centered around verbs \cite{YatskarZF16, CoorayCL20, PrattYWFK20, YuWLXG23} . This framework identifies the objects involved in the activities, as well as the roles these objects play within the activities.
Cooray et al. \cite{CoorayCL20} propose treating the prediction of semantic roles as a query-based visual reasoning problem. 
Pratt et al. \cite{PrattYWFK20} introduce the joint situation localizer, which aims to simultaneously predict semantic frames and groundings. 
Yu et al. \cite{YuWLXG23} leverage diverse statistical knowledge to enable neural networks to perform adaptive global reasoning on nouns.

\section{Method}

In this section, we first give an overview of using the news-aware semantic rule to generate image descriptions. 
Then, we delineate the computational pipeline of Named Entity Extraction (see Section \ref{Named Entity Extraction}), News Rule Construction (see Section \ref{News Rule Construction}), and Caption Generation (see Section \ref{Caption Generation}). 

\subsection {Problem Formulation}
Given an image $I$ and its corresponding news article $A$, the objective of the news captioning task is to generate a coherent natural language description $Y=\{y_1,y_2,\dots,y_T\}$, 
where $y_i$ denotes the $i$-th word; $T$ is the length of the sentence. 
The whole process can be formulated as: 
\begin{equation}\label{1}
    \log p(Y|I,A;\theta)=\sum_{t=1}^T\log p(y_t|I,A,y_{0:t-1};\theta)
\end{equation}
where $p(\cdot)$ is a probability distribution function; $y_{0:t-1}=\{y_0,y_1,\dots,y_{t-1}\}$ is the series of words produced before the time step $t$; $\theta$ represents the trainable parameters of the model. 

In order to generate captions that adhere to fundamental rules of news reporting, we first design the news-aware semantic rule $F$, which is filled using images and news articles. Then, 
we integrate this semantic rule into the large-scale pre-trained model (e.g., BART), where multiple encoder layers are embedded with news-aware semantic rule, to generate image descriptions. Formally, Eq. \ref{1} is re-written as: 
\begin{equation}\label{2}
\log p(Y|F,I,A;\theta) = \sum_{t=1}^T \log p(y_t|F^{(M)},I,A,y_{0:t-1};\theta)
\end{equation}
where $M$ represents the number of encoder layers with embedded rules.

\begin{figure*}[t]
  \begin{center}
  \includegraphics[width=\linewidth]{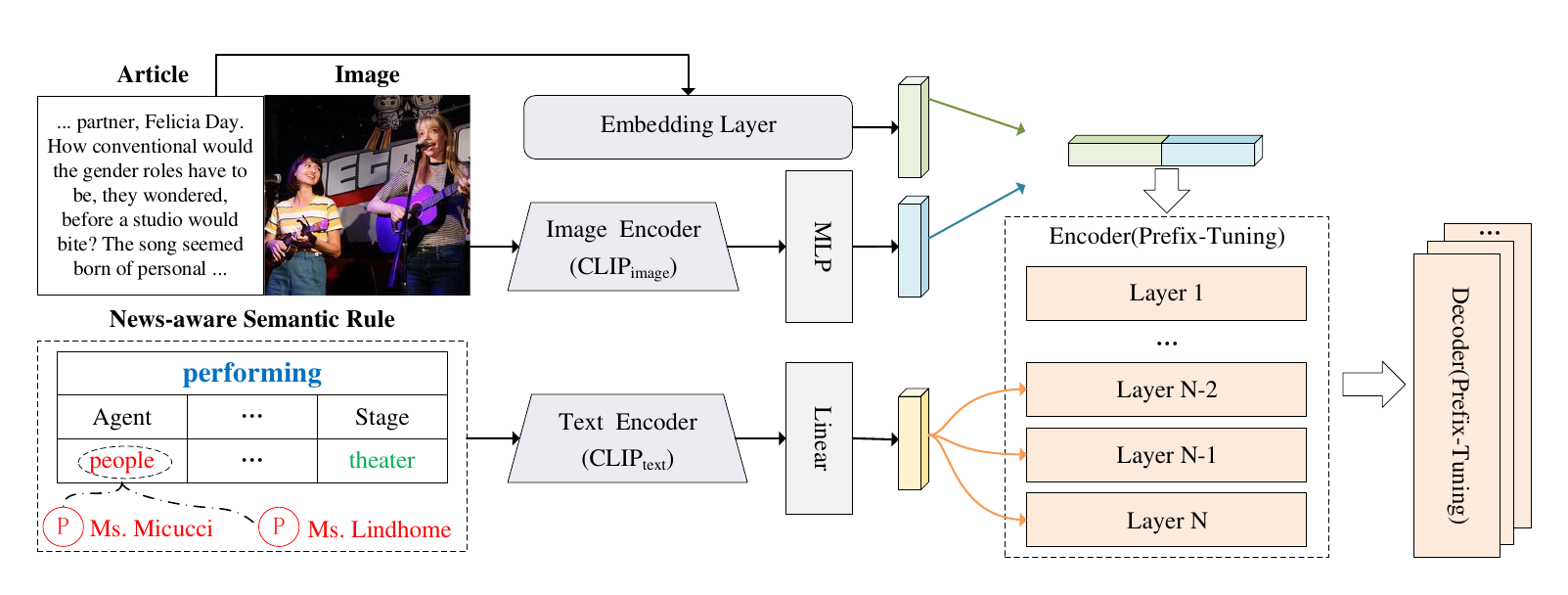}\\
  \caption{The detailed process of the caption generation. In order to ensure that BART effectively follows the specified rule signal, we integrate the news-aware semantic rule into the last three encoder layers of BART, using the prefix-tuning strategy. }\label{detail_diagram}
  \end{center}
\end{figure*}

\subsection {Named Entity Extraction} \label{Named Entity Extraction}
Named entities play a crucial role in enhancing the accuracy and richness of news captions by providing specific person's name (PER), organization (ORG), and location (LOC). 
In this section, we extract named entities and their corresponding types from given news articles, which are then used in subsequent modules. 
We first employ a pre-trained named entity recognition (NER) model \footnote{https://huggingface.co/Davlan/distilbert-base-multilingual-cased-ner-hrl\label{NER}} to extract a set of entities $E=\{e_1,e_2,\dots,e_L\}$ from the article $A$, where $L$ is the number of named entities.
As shown in Fig. \ref{model_diagram}(a), to select the named entities that closely align with the news image content, we use the CLIP model \cite{RadfordKHRGASAM21} to evaluate the similarity between the image and each entity.
However, due to significant discrepancies between the dataset used in the pre-training of CLIP and the data from the news domain, we fine-tune the CLIP model on two existing news caption datasets separately (i.e., GoodNews \cite{BitenGRK19} and NYTimes800k \cite{TranMX20}).
Specifically, we define a correct image-text pair as a news image with its matching caption. Incorrect captions serve as negative samples. The CLIP model uses a contrastive loss function to increase the similarity of matching pairs and decrease it for mismatched pairs, enhancing the model's skill in accurately linking images and text in news domain.
The fine-tuned CLIP model is used to compute the similarity scores between the image $I$ and each entity $e_i\in E$, and then select the top-$k$ entities to build the new entity set $E_k$: 

\begin{equation}\label{3}
E_k =\{e_i\}_{i=1}^k= \mathrm{arg~Top\mbox{-}k}(\mathrm{CLIP_{news}}(I,E))
\end{equation}
where $\mathrm{CLIP_{news}} (I, E)$ aims to calculate image-text matching scores by the fine-tuned CLIP model \cite{RadfordKHRGASAM21}.

\subsection {News Rule Construction} \label{News Rule Construction}
We design the news-aware semantic rule for the sentence generation. 
As shown in Fig. \ref{model_diagram}(b), given an image $I$, we first utilize the advanced
situation recognition technique \cite{PrattYWFK20} to extract a critical verb that can represent the primary activity depicted in the image (e.g., performing). 
Based on this verb, we employ the verb dictionary, such as FrameNet \cite{BakerFL98}, to retrieve a collection of semantic roles associated with it, including ``Agent'' and ``Stage''. 
These semantic roles categorize the involvement of objects in this activity, allowing for a detailed segmentation and comprehension of the context surrounding the critical verb.
Then, we use the situation recognition technique to predict and correlate generic objects with their corresponding role, such as ``people'' with ``Agent'', ``theater'' with ``Stage''. 
Formally, we represent the critical verb, semantic roles, and objects as $F'$:
\begin{equation}\label{4}
F'=\{v_0;(r_1,s^{(r_1)}),(r_2,s^{(r_2)}),\dots,(r_n,s^{(r_n)})\}
\end{equation}
where $v_0$ represents the verb; $r_i$ refers to the semantic role correlated with the verb; $s^{(r_i)}$ refers to the object corresponding to the semantic role $r_i$; $n$ is the number of semantic roles.

\begin{algorithm}[t]
\SetKwData{Left}{left}\SetKwData{This}{this}\SetKwData{Up}{up}
\SetKwFunction{Union}{Union}\SetKwFunction{FindCompress}{FindCompress}
\SetKwInOut{Input}{input}\SetKwInOut{Output}{output}
\caption{Solution of Named Entity Replacement}\label{alg:algorithm_1}
\Input{$F'$, $E_k^p$, $D$}
\Output{$F$}
\BlankLine
  \For {$j\leftarrow 1$ \KwTo $n$}{
  \Repeat {line search criterion is satisfied}{
    \If{$s^{(r_j)}$ belong to $D$}{
        $(r_j,s^{(r_j)}) \gets (r_j,E_k^p)$\;
        }
     }
    \If {the objective stop criterion satisfied}{
    update $F'$ to $F$\;
    \Return $F$
    }
  }
\end{algorithm}

Next, we select named entities from news articles to replace the generic objects in $F'$. 
We first manually construct a vocabulary D, which includes a range of generic objects that can represent PER, ORG, and LOC (e.g., objects like ``people", ``man", and ``woman" can represent PER). We then categorize the named entities in $E_k$ (Eq. \ref{3}) into three subsets $E_k^p$, $E_k^o$, and $E_k^l$ with respect to three types, i.e., PER, ORG, and LOC, respectively. 
During the traversal of $F'$, if we find an object of the vocabulary D that exists in $F'$, we will replace this object in $F'$ with named entities from the corresponding type subset. 
As shown in Fig. \ref{model_diagram}(b), we replace the generic object ``people'' in $F'$ with the named entities ``Ms. Micucci” and ``Ms. Lindhome”. It can be formulated as follows: 
\begin{equation}\label{5}
(r_j,s^{(r_j)}) \gets (r_j,E_k^p),  \text{where~} s^{(r_j)} \in D
\end{equation}
where $(r_j,s^{(r_j)})$ is the $j$-th role-object pair in $F'$.
The procedure is summarized in Algorithm \ref{alg:algorithm_1} (take PER as an example). 

Till now, we have obtained the news-aware semantic rule representation, denoted as $F$, which presents a clear affiliation between generic objects in images and named entities in news articles, and includes the set of semantic roles surrounding the primary action. 

\subsection {Caption Generation} \label{Caption Generation}
News captioning task can be understood as generating accurate and concise summaries by integrating images and news articles. BART is a large-scale pre-trained summarization model \cite{LewisLGGMLSZ20}, which is composed of a bidirectional encoder and an autoregressive decoder. It exhibits excellent performance in various text generation tasks \cite{liu2021kg}, \cite{li2022contrast}.
In this paper, we employ the news-aware semantic rule to guide BART for the news caption generation. 
Fig. \ref{detail_diagram} illustrates the detailed process of the caption generation. To ensure the effective adherence of BART to the designated rule signal, we incorporate the semantic rule into multiple encoder layers using the prefix-tuning strategy. 

We employ the frozen CLIP model to encode the image $I$ and the semantic rule $F$ independently, and obtain the visual features and the rule features, respectively. 
The former is embedded into the visual vector $z^v$ by a multi-layer perceptron (MLP), while the latter is transformed into the rule vector $z^f$ through a linear layer:
% \begin{equation}\label{6}
% v=\mathrm{MLP}(\mathrm{CLIP_{image}}(I))
% \end{equation}
% \begin{equation}\label{7}
% f=\mathrm{Linear}(\mathrm{CLIP_{text}}(F))
% \end{equation}
\begin{equation}\label{6}
z^v = \mathrm{MLP}(\mathrm{CLIP_{image}}(I))
\end{equation}
\begin{equation}\label{7}
z^f=\mathrm{Linear}(\mathrm{CLIP_{text}}(F))
\end{equation}
where $\mathrm{CLIP_{image}}(\cdot)$ is the image encoder; $\mathrm{CLIP_{text}}(\cdot)$ is the text encoder; 
For the article representation, we use the tokenizer to convert the news article into a collection of tokens $X^A=\{x_1,x_2,\dots,x_N\}$, where $x_i$ represents the $i$-th token in the article. We feed $X^A$ into an embedding layer and obtain the article vector $x^A$.

Essentially, the news captioning task requires adhering to some fundamental rules of news reporting, which is significantly different from the previous BART objective for the general summarization task.
In this section, we utilize the prefix-tuning strategy \cite{LiL20}. This lightweight fine-tuning strategy involves optimizing a tiny task-specific vector called ``prefix", which can inject the task-specific structural information during the fine-tuning process. Therefore, it can help the model better adapt to the requirements of the news captioning task. 
First, we concatenate the rule vector $z^f$, the visual vector $z^v$, and the article vector $x^A$, which are input into the BART encoder: 
\begin{equation}\label{8}
c=\mathrm{BART_{enc}}(z^f,z^v,x^A)
\end{equation}
where $c$ is the output vector from the BART encoder; $\mathrm{BART_{enc}}(\cdot)$ is the bidirectional encoder of BART.
 
Specially, we inject the news-aware semantic rule vector $z^f$ into the last three encoder layers of BART (i.e., $M$ is set as 3 in Eq. \ref{2}). 
It is because compared with shallow encoder layers, deep ones (last three layers) are closer to the prediction outputs and have better semantic modeling capabilities.
Therefore, embedding the news-aware semantic rule into the deep layers can boost the captioner to generate image descriptions following designated rule signal. 
It can be formulated as: 
\begin{equation}\label{9}
\begin{split}
    X^{(l+1)} =& \mathcal{F}(\mathrm{LayerNorm}( \\
    X^{(l)}+&\mathrm{MultiHead}([z^f;X^{(l)}],[z^f;X^{(l)}],[z^f;X^{(l)}])))
\end{split}
\end{equation}
\begin{equation}\label{10}
\mathrm{MultiHead}(Q,K,V)=\mathrm{Concat}(h_1,h_2,...,h_n)W^O
\end{equation}
\begin{equation}\label{11}
h_i=\mathrm{Attention}(Q W_i^Q,K W_i^K,V W_i^V),
\end{equation}
\begin{equation}\label{12}
\mathrm{Attention}(Q,K,V)=\mathrm{Softmax}(\frac{Q[p;K]^T}{\sqrt{d}})[p;V]
\end{equation}
where $X^{(l+1)}$ and $X^{(l)}$ are the input and output of the current encoder layer, respectively; 
$\mathcal{F}$ is the feed-forward layer; $\mathrm{LayerNorm}(\cdot)$ is the layer normalization; 
$z^f$ is the rule vector; 
$[;]$ represents the concatenation operation; 
$n$ is the number of parallel attention layers; 
$W_i^Q$, $W_i^K$, $W_i^V$, and $W^O$ are learnable weight matrices; 
$p$ is the prefix vector of the encoder.

Given the representation $c$ from the BART encoder (Eq. \ref{8}) and the partial text sequence $y_{0:t-1}$  generated before the time step $t$, the BART decoder iteratively generates the next word to eventually produce a complete image description $Y=\{y_1,y_2,\dots,y_T\}$:
\begin{equation}\label{13}
y_t\sim p(y_t|c,y_{0:t-1})=\mathrm{BART_{dec}}(c, y_{0:t-1})
\end{equation}
where $y_t$ is the $t$-th word in the generated caption; $\mathrm{BART_{dec}}(\cdot)$ is the autoregressive decoder of BART.

\section{Experiment}
% \usepackage{booktabs}
% \usepackage{multirow}
% \usepackage{graphicx}
\begin{table*}[!t]\small\renewcommand\arraystretch{1.0}
\centering
\caption{EVALUATION RESULTS ON GOODNEWS AND NYTIMES800K DATASETS. ALL VALUES ARE REPORTED AS PERCENTAGES. }
\label{tab:experiment}
\scalebox{1}
% \resizebox{\columnwidth}{}
{%
\begin{tabular}{@{}cl|cccc|cccccc@{}}
\toprule
\multirow{2}{*}{\textbf{}} &
  \multirow{2}{*}{\textbf{Method}} &
  \multirow{2}{*}{\textbf{BLEU-4}} &
  \multirow{2}{*}{\textbf{METEOR}} &
  \multirow{2}{*}{\textbf{ROUGE}} &
  \multirow{2}{*}{\textbf{CIDEr}} &
  \multicolumn{2}{c}{\textbf{Named Entities}} &
  \multicolumn{2}{c}{\textbf{People’s Names}}  &
  \multicolumn{2}{c}{\textbf{Rare Proper Nouns}} \\
                              &      &      &       &       &       & \textbf{P}       & \textbf{R}     & \textbf{P}     & \textbf{R}   & \textbf{P}     & \textbf{R}\\ \midrule
\multirow{11}{*}
{\rotatebox{90}{GoodNews}}    & SAT(2015)\cite{XuBKCCSZB15}             & 0.73 & 4.14  & 11.88 & 12.15 & 8.19 & 7.10  & -   & -   & -   & -    \\
                              & Att2in2(2017)\cite{RennieMMRG17}        & 0.76 & 3.90  & 11.58 & 11.58 & - & - & - & - & -   & -           \\
                              & Ramisa et al.(2019)\cite{RamisaYMM18}   & 0.89 & 4.45  & 12.09 & 15.35 & - & -   & -   & -   & -   & -        \\
                              & Biten et al.(2019)\cite{BitenGRK19}     & 0.89 & 4.37  & 12.20 & 13.10 & 8.23 & 6.06  & 9.38   & 6.55   & 1.06 & 12.50     \\
                              & ICECAP(2020)\cite{abs-2108-02050}       & 1.96 & 6.01  & 15.70 & 26.08 & -   & -  & -  & -   & -  & -     \\
                              & Tell(2020)\cite{TranMX20}               & 6.05 & 10.30 & 21.40 & 53.80 & 22.20   & 18.70  & 29.20  & 23.10  & 15.60 & 26.30  \\
                              & Zhao et al.(2021)\cite{abs-2107-11970}  & 6.14 & 6.32  & 21.46 & 54.02 & -      & -    & -   & -  & -   & - \\
                              & JoGANIC(2021)\cite{YangKTJ21}           & 6.83 & 11.25 & 23.05 & 61.22 & \textbf{26.87}     & 22.05         & -     & -   & -          & -   \\
                              & Zhou et al.(2022)\cite{ZhouLRY22}       & 6.30 & -     & 22.40 & 60.30 & 24.20              & 20.90         & -    & -   & -    & -   \\ 
                              & NewsMEP(2022)\cite{ZhangFMZ022}         & \textbf{8.30} & 12.23 & 23.17   & 63.99          & 23.43      & 23.24    & 32.25   & 29.06  & -   & -          \\ \cmidrule(l){2-12}
                              & \textbf{Ours}                   & 8.18 & \textbf{12.50} & \textbf{23.56} & \textbf{71.58} & 25.51    & \textbf{23.68}    & \textbf{33.89}    & \textbf{30.09} &  \textbf{21.50} &    \textbf{37.53}        \\ \midrule
% \multicolumn{1}{l}{\multirow{5}{*}{\rotatebox{90}{NYTimes800K}}} & Tell (2020)        & 6.30 & 10.30 & 21.70 & 54.40 & 24.6                 & 24.6                 \\
\multirow{6}{*}
{\rotatebox{90}{NYTimes800k}} & Tell (2020)\cite{TranMX20}              & 6.30 & 10.30 & 21.70 & 54.40 & 24.60    & 22.20     & 37.30   & 31.10   & 34.20 & 27.00     \\
                              & Zhao et al.(2021)\cite{abs-2107-11970}  & 6.32 & 6.25  & 21.62 & 54.47 & -     & -  & -  & -       & -      & -   \\
                              & JoGANIC(2021)\cite{YangKTJ21}  & 6.79 & 10.93 & 22.80 & 59.42 & 28.63     & 24.49         & -  & -  & -    & -  \\
                              & Zhou et al.(2022)\cite{ZhouLRY22}       & 7.00 & -     & 22.90 & 63.60 & \textbf{29.80}  & 25.90         & -    & -  & -      & -   \\ 
                              & NewsMEP(2022)\cite{ZhangFMZ022}         & \textbf{9.57} & 13.02 & 23.62 & 65.85 & 26.61  & 28.57   & 41.32  & 38.51 & -  & -       \\ \cmidrule(l){2-12}
                              & \textbf{Ours}                           & 9.41 & \textbf{13.10} & \textbf{24.42} & \textbf{72.29} & 28.15 & \textbf{28.80}  & \textbf{41.46} & \textbf{39.88} & \textbf{40.87} & \textbf{38.36}               \\ \bottomrule
\end{tabular}%
}
\end{table*}
\subsection {Experimental Setup}
\noindent \textbf {Datasets.} 
The proposed method is evaluated on two large-scale news image captioning datasets: GoodNews \cite{BitenGRK19} and NYTimes800k \cite{TranMX20}, both of which are sourced from the New York Times. The GoodNews dataset, which comprises 257,033 articles and 462,642 images, provides 421K captions for training, 18K captions for validation, and 23K captions for testing. In contrast, the NYTimes800k dataset exhibits greater scale. It consists of 763K training, 8K validation, and 22K test captions, along with a total of 444,914 articles and 792,971 images. Notably, each article in the NYTimes800k dataset may correspond to multiple images.

\noindent \textbf {Metrics.} 
To evaluate the generated captions, we employ general caption generation metrics, namely BLEU \cite{PapineniRWZ02}, ROUGE \cite{lin-2004-rouge}, METEOR \cite{DenkowskiL14}, and CIDEr \cite{VedantamZP15}. 
BLEU and ROUGE evaluate the accuracy and coverage of the generated descriptions, while METEOR assesses their fluency and grammatical correctness. 
Comparatively, CIDEr places greater emphasis on rare words, thereby yielding more persuasive evaluation results for news image captioning models. 
Furthermore, we provide Precision (P) and Recall (R) scores for named entities, people's names and rare proper nouns. We use the spaCy toolkit \footnote{https://spacy.io/ \label{spaCy}} to recognize named entities in both the generated sentences and the ground-truth sentences.

\noindent \textbf {Training Details.} 
The BART \footnote{https://huggingface.co/facebook/bart-large \label{BART}} model comprises 12 encoder layers and 12 decoder layers, with each self-attention block incorporating 16 heads. The maximum length of input news text is set to 512 tokens, and each token has an embedding dimension of 1024. We assign a maximum length of 60 to the generated captions for the training and validation sets and 128 for the test set. 
In Eq. 3, k is set to 3. 
Considering that the maximum input length of pre-trained named entity recognition model is 512 tokens, we perform slicing operations on the original news articles to ensure the model's capability to handle the entire article. 
We employ the AdamW optimizer \cite{LoshchilovH19} and a linear learning rate scheduler during the training process of BART. We set the batch size to 16, the training period to 30 epochs, and the learning rate to $5 \times 10^{-5}$. The work is completed using the PyTorch framework, and the entire model is trained on two RTX 3090Ti GPUs. The training time per epoch is approximately 1.5 hours for the GoodNews dataset and approximately 3 hours for the NYTimes800k dataset.
% \textcolor{blue}{We fine-tuned CLIP on the training sets of GoodNews and NYTimes800k datasets, respectively. The training data for Goodnews includes 421K image-caption pairs, while NYTimes800k includes 763K image-caption pairs. The learning rate is $5 \times 10^{-6}$. The batch size is set to 50. The CLIP is trained on an RTX 3090Ti GPU for 2 hours for GoodNews and 3 hours for NYTimes800k.}
Besides, we fine-tune the CLIP using 421K image-caption pairs from GoodNews and 763K pairs from NYTimes800k. 
The learning rate is $5 \times 10^{-6}$. The batch size is 50. The CLIP is fine-tuned on one RTX 3090Ti GPU for 2 hours on GoodNews and 3 hours on NYTimes800k.

\subsection {Comparison with the State-of-The-Art}

% table_3
\begin{table*}[!t]\small\renewcommand\arraystretch{1.0}
\centering
\caption{RESULTS OF ABLATION STUDIES ON THE PROPOSED METHOD ON GOODNEWS AND NYTIMES800K DATASETS.}
\label{tab:ablation_1}
\scalebox{1}
{%
\begin{tabular}{@{}cl|cccc|cccccc@{}}
\toprule
\multirow{2}{*}{\textbf{}} &
  \multirow{2}{*}{\textbf{Method}} &
  \multirow{2}{*}{\textbf{BLEU-4}} &
  \multirow{2}{*}{\textbf{METEOR}} &
  \multirow{2}{*}{\textbf{ROUGE}} &
  \multirow{2}{*}{\textbf{CIDEr}} &
  \multicolumn{2}{c}{\textbf{Named Entities}} &
  \multicolumn{2}{c}{\textbf{People’s Names}} &
  \multicolumn{2}{c}{\textbf{Rare Proper Nouns}}\\
                              &                  &      &       &       &       & \textbf{P}    & \textbf{R}     & \textbf{P}     & \textbf{R}   & \textbf{P}     & \textbf{R}\\ \midrule
\multirow{4}{*}
{\rotatebox{90}{GoodNews}}    & Non-Rule         & 7.62 & 11.22  & 22.02 & 65.37 & 23.56          & 21.67          & 32.48          & 28.02          & 20.39          & 37.33\\
                              & Non-Entity       & 7.93 & 11.60  & 22.78 & 68.41 & 24.34          & 22.45          & 33.04          & 28.80          & 20.76          & 37.19\\
                              & PER-Rule      & 7.97 & 11.62  & 22.82 & 69.12 & 24.25          & 22.28          & 33.22          & 28.82          & 20.61          & 37.26\\ \cmidrule(l){2-12}
                              & \textbf{Ours}              & \textbf{8.18} & \textbf{12.50} & \textbf{23.56} & \textbf{71.58} & \textbf{25.51}    & \textbf{23.68}    & \textbf{33.89}    & \textbf{30.09} &  \textbf{21.50} &    \textbf{37.53}     \\
                               \midrule
\multirow{4}{*}
{\rotatebox{90}{NYTimes}}     & Non-Rule         & 9.26 & 12.42  & 23.11 & 69.74  & 26.17 & 26.45 & 39.48 & 36.58  & 39.89 & 37.12       \\
                              & Non-Entity       & 9.21 & 12.41  & 23.48 & 72.15 & 27.04 & 26.53 & 40.53 & 37.30 & 40.52 & 37.63  \\
                              & PER-Rule      & 9.37 & 12.52  & 23.43 & 71.63  & 27.02 & 26.57 & 40.55 & 37.54 & 40.03 & 37.25   \\ \cmidrule(l){2-12}
                              & \textbf{Ours}             & \textbf{9.41} & \textbf{13.10} & \textbf{24.42} & \textbf{72.29} & \textbf{28.15} & \textbf{28.80}  & \textbf{41.46} & \textbf{39.88} & \textbf{40.87} & \textbf{38.36}     \\
                               \midrule

\end{tabular}%
}
\end{table*}
% table_3

We compare the proposed method with the state-of-the-art methods, which are divided into two groups. 
1) \textbf{Template-based methods.} 
It first generates template captions with placeholders, and then selects appropriate named entities from news articles to replace these placeholders to produce captions \cite{BitenGRK19}, \cite{RamisaYMM18}, \cite{XuBKCCSZB15}, \cite{RennieMMRG17}. 
For example, Biten et al. \cite{BitenGRK19} use named entity types (e.g., ``PER", ``ORG") as placeholders and select named entities via the highest attention value to fill placeholders. 
2) \textbf{End-to-end methods.} 
It directly integrates news images and articles to generate the named entity-aware descriptions \cite{TranMX20}, \cite{YangKTJ21}, \cite{ZhangFMZ022}, \cite{abs-2108-02050}, \cite{abs-2107-11970}, \cite{ZhouLRY22}. 
For examples, Tell \cite{TranMX20} uses the byte-pair encoding to generate unseen or rare named entities.
NewsMEP \cite{ZhangFMZ022} uses visual and entity-oriented prompts to fine-tune a large-scale pre-trained model. 

Tab. \ref{tab:experiment} shows that our method can achieve competitive performance against the state-of-the-arts on both GoodNews and NYTimes800k datasets. We have the following three main observations:

\begin{itemize}

\item Our method can significantly outperform all template-based methods and achieve competitive performance compared with end-to-end methods. 
Specifically, the template-based methods are limited by the capabilities of the template generator, which can not generate descriptions with diverse and expressive language. 
The end-to-end methods overlook the guidance of news-aware rule signal, resulting in the generated sentences that may not adhere to the fundamental rules of news reporting. 
Comparatively, our method builds the news-aware semantic rule to guide the model to generate news captions, which can accurately describe the individuals and actions associated with the event. 
As shown in Tab. \ref{tab:experiment}, on GoodNews, our method can outperform the best template-based competitor \cite{BitenGRK19} by 58.48\%, and the best end-to-end competitor \cite{ZhangFMZ022} by 7.59\% under CIDEr scores, respectively. 

% \item \textcolor{blue}{Tab. \ref{tab:experiment} shows our method can explicitly describe people's names, and rare proper nouns against state-of-the-art methods. 
% Thanks to the proposed news-aware semantic rule, which can adhere to the fundamental rules of news reporting, our method can achieve impressive results of 36.99\% and 37.78\% on GoodNews and NYTimes800k, respectively, under the recall of rare proper nouns. 
% It makes the absolute improvement over the best competitor Tell \cite{TranMX20} by 10.69\% and 10.78\%. 
% However, we have observed that JoGANIC \cite{YangKTJ21} and Zhou et al. \cite{ZhouLRY22} slightly outperform our method in terms of the precision score for named entities. It is because both of them benefits most from using the richer named entity information. In contrast, we focus solely on three types of named entities: person's name (PER), organization (ORG), and location (LOC). }
\item Tab. \ref{tab:experiment} shows our method can explicitly describe named entities, people's names, and rare proper nouns against state-of-the-art methods. 
Thanks to the proposed news-aware semantic rule, which can adhere to the fundamental rules of news reporting, our method can achieve impressive results of 37.53\% and 38.36\% on GoodNews and NYTimes800k, respectively, under the recall of rare proper nouns. 
It makes the absolute improvement over the best competitor Tell \cite{TranMX20} by 11.23\% and 11.36\%. 
However, we have observed that JoGANIC \cite{YangKTJ21} and Zhou et al. \cite{ZhouLRY22} slightly outperform our method in terms of the precision score for named entities. It is because both of them benefit most from using the richer named entity information. In contrast, we focus solely on three types of named entities: person's name (PER), organization (ORG), and location (LOC).

\item  NewsMEP \cite{ZhangFMZ022} also uses the pre-trained large-scale model to generate news captions. 
Comparatively, we use the news-aware semantic rule to guide the large-scale model, which can generate sentences that adhere to the fundamental rules of news reporting. 
The improved results in terms of METEOR, ROUGE, and CIDEr illustrate our method can produce more human-like and fluent sentences compared to NewsMEP. Furthermore, 
our method outperforms NewsMEP in describing named entities and people's names, highlighting our  ability to accurately capture the named entities surrounding news content.

\end{itemize}

\subsection {Ablative Analysis}\label{Ablative Analysis}

To examine the efficacy of the proposed method, we design a series of comprehensive experiments by answering the following questions. 
\textbf{Q1}: Is the strategy of using the news-aware semantic rule effective? 
\textbf{Q2}: Is it necessary to embed named entities into the semantic rule? 
\textbf{Q3}: Is there another choice to compute the semantic rule? 
\textbf{Q4}: Which layers should be selected in BART to embed semantic rules? 

\noindent \textbf{Effectiveness of Semantic Rule (Q1).} 
In this paper, we use the news-aware semantic rule, which consists of the primary action depicted in an image and the roles played by named entities, to guide the image caption generation. 
To validate the effectiveness of this rule, we design the variant Non-Rule, which removes the processes of the named entity extraction and the news rule construction from our full model. 
Specifically, we directly feed the news image and the corresponding article content to the BART model. Our method degenerates to Eq. \ref{1} and Eq. \ref{8} is rewritten as: 
\begin{equation}\label{14}
c=\mathrm{BART_{enc}}(z^v,x^A)
\end{equation}
As shown in Tab. \ref{tab:ablation_1}, our method performs better than Non-Rule across all metrics on two datasets. 
It illustrates the proposed rule-driven method can facilitate 
the model to accurately describe the individuals and actions associated with the event, making the output sentences more in line with the fundamental rules of news reporting. 

% \input{tab_ablation_3}
% \usepackage{booktabs}
% \usepackage{multirow}
% \usepackage{graphicx}
\begin{table*}[!t]\small\renewcommand\arraystretch{1.0}
\centering
\caption{Performance of the proposed method on GoodNews and NYTimes800k datasets when using different BART layers to embed the news-aware semantic rule. }
\label{tab:ablation_2}
\scalebox{1}
% \resizebox{\columnwidth}{}
{%
\begin{tabular}{@{}cc|cccc@{}}
\toprule
                       & \textbf{Encoder Layer}       & \textbf{BLEU-4}& \textbf{METEOR}& \textbf{ROUGE}& \textbf{CIDEr}     \\ \midrule
\multirow{4}{*}{GoodNews}    & P1(1 $\sim$ 3)         & 7.66          & 11.34          & 22.44          & 66.76                \\
                             & P2(4 $\sim$ 6)         & 8.01          & 11.68          & 22.98          & 69.17                \\
                             & P3(7 $\sim$ 9)         & 7.97          & 11.67          & 22.96          & 68.82                \\
                             & P4(10 $\sim$ 12)       & \textbf{8.18} & \textbf{12.50} & \textbf{23.56} & \textbf{71.58}       \\ \midrule
\multirow{4}{*}{NYTimes800k} & P1(1 $\sim$ 3)         & 9.17          & 12.36          & 23.08          & 70.01               \\
                             & P2(4 $\sim$ 6)         & 9.21          & 12.36          & 23.14          & 70.09                \\
                             & P3(7 $\sim$ 9)         & 9.36          & 12.55          & 23.50          & 71.77                \\
                             & P4(10 $\sim$ 12)       & \textbf{9.41} & \textbf{13.10} & \textbf{24.42} & \textbf{72.29}     \\
                             % & \multicolumn{1}{l}{} & \multicolumn{1}{l}{} & \multicolumn{1}{l}{} & \multicolumn{1}{l}{} & \multicolumn{1}{l}{} \\ 
                             \bottomrule
\end{tabular}
}
\end{table*}
\noindent \textbf{Effectiveness of Embedding Named Entities (Q2). }
During the rule construction, we employ named entities extracted from news articles to replace generic objects in Algorithm \ref{alg:algorithm_1}. 
In order to verify its effectiveness, we design the variant Non-Entity, which removes the process of named entity extraction and replacement (Eq. \ref{5}) from our full model. 
Namely, we directly use $F'$ in Eq. \ref{4} as the rule to guide the model learning. 
We encode $F'$ using CLIP and transform it into the rule vector $z^{f'}$ through a linear layer. We rewrite Eq. \ref{7} as: 
\begin{equation}\label{15}
z^{f'}=\mathrm{Linear}(\mathrm{CLIP_{text}}(F'))
\end{equation}
As shown in Tab. \ref{tab:ablation_1}, our method shows superior performance on all evaluation metrics. This verifies that clarifying the affiliation between generic objects in images and named entities in news articles can enhance the model's ability to accurately describe named entities. 

\noindent \textbf{Investigation on News Rule (Q3).} 
\begin{figure}[t]
  \begin{center}
  \includegraphics[width=0.8\linewidth]{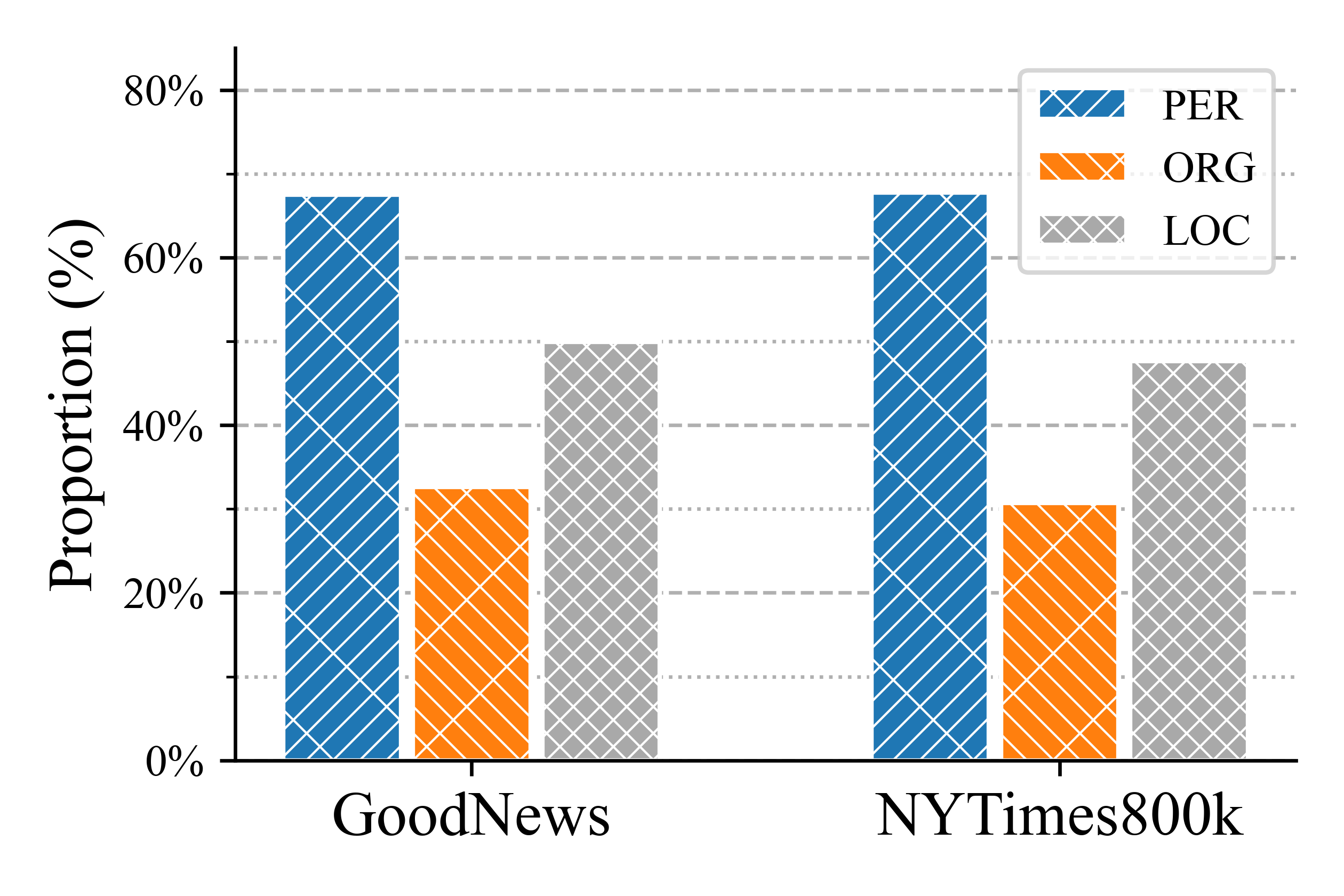}\\
  \caption{The proportion of three types of named entities, i.e., PER (person's name), ORG (organization), and LOC (location), in the training corpus of GoodNews and NYTimes800k.}\label{statistics_diagram}
  \end{center}
\end{figure}
Our news-aware semantic rule covers person's name (PER), organization (ORG), and location (LOC) that are related to the primary action in images. 
After conducting statistical analysis on datasets, we have discovered that the frequency of PER is significantly higher compared to other categories. 
As shown in Fig. \ref{statistics_diagram}, the GoodNews dataset and the NYTimes800k dataset have 67.50\% and 67.75\% of training captions, respectively, which include at least one person's name. 
To verify the effectiveness of our semantic rule, in this section, we design another rule strategy, which mainly focuses on the people's names related to primary actions. 
Specifically,
we design the variant PER-Rule that only uses the named entities of the PER category to form the rule to guide the BART. 

As shown in Tab. \ref{tab:ablation_1}, the variant PER-Rule performs worse than our method on both datasets. This result proves that our news-aware semantic rule has superior applicability and reliability in news reporting scenarios. 

\noindent \textbf{Investigation on Rule-embedded Layers (Q4). }
We embed the constructed semantic rules into the large-scale pre-trained model, BART. 
Particularly, the BART model includes 12 encoder layers, each with the ability to embed semantic information of different complexities. 
We argue that deep layers are close to the prediction outputs, which have better semantic modeling capabilities. 
Therefore, we embed the news-aware semantic rules into the last three encoder layers to guide BART for the caption generation. 
In this section, we evaluate the performance of other layers to embed rules. 
As shown in Tab. \ref{tab:ablation_2}, we divide the 12 encoder layers into four sequential parts, with each part consisting of three adjacent encoder layers, i.e., P1(1 $\sim$ 3); P2(4 $\sim$ 6); P3(7 $\sim$ 9); P4(10 $\sim$ 12). 

We observe that the deep layers P4 can achieve the better performances than the shallow layers P1 or P2, across all metrics on two datasets. 
Especially, in the CIDEr score, our method (P4) can achieve 71.58\% on GoodNews and 72.29\% on NYTimes800k, with a relative improvement of 4.82\% and 2.28\% over P1. 
It illustrates that the deeper encoder layers are more suitable for embedding news-aware semantic rules in image caption generation. 

\subsection {Qualitative Analysis}

\begin{figure*}[ht!]
\centering
\includegraphics[width=\linewidth]{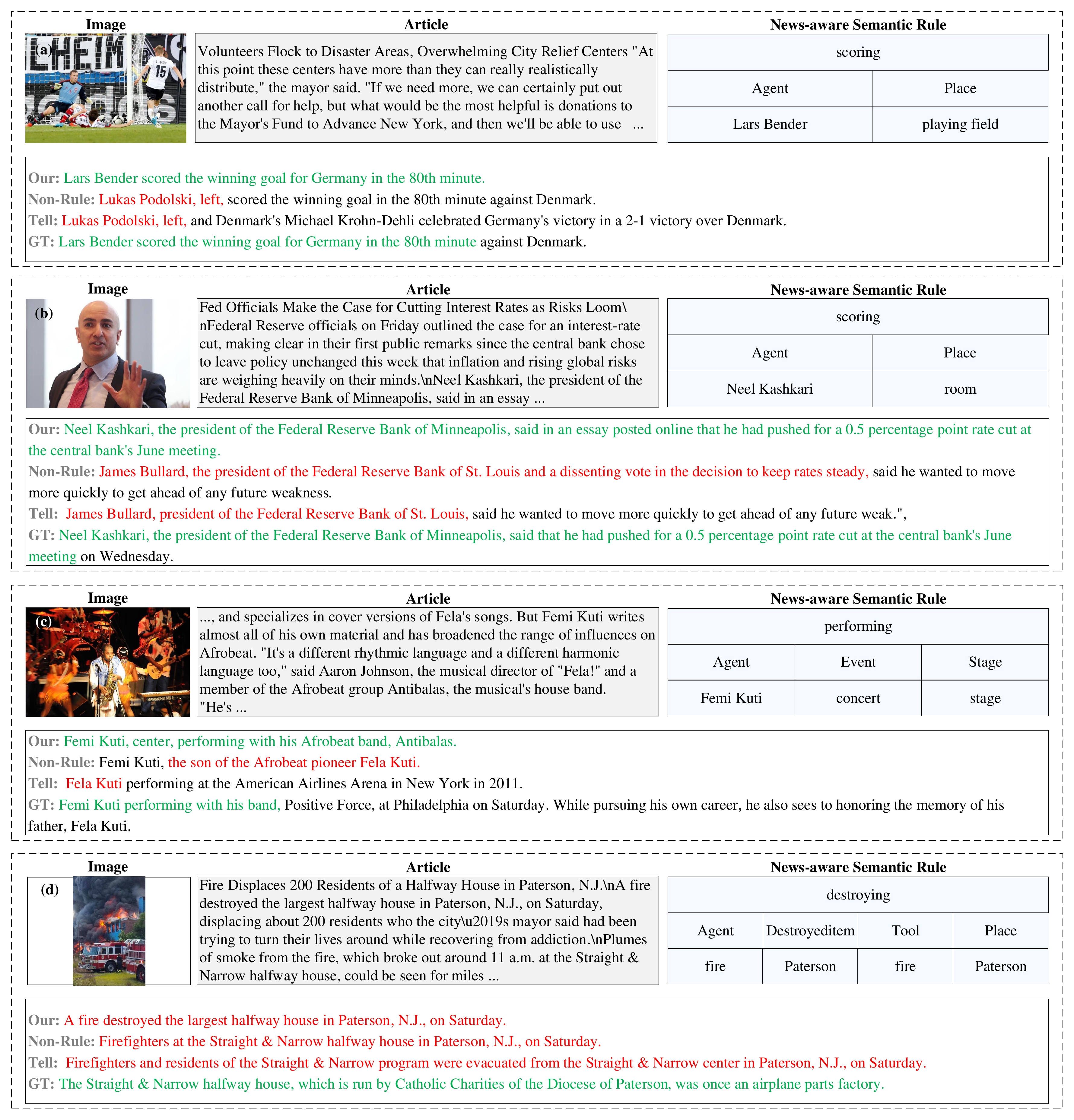}
\caption{Qualitative results of our method. We provide the constructed news-aware semantic rule in each example. \textbf{Non-Rule} refers to the caption generated without using news-aware semantic rule. \textbf{Tell} uses the byte-pair-encoding transformer to generate captions \cite{TranMX20}. \textbf{GT} is the ground-truth caption. Correct sentences are marked in \textcolor[RGB]{0, 176, 80}{green}, while incorrect ones are marked in \textcolor{red}{red}. }
\label{visual_diagram}
\end{figure*}

Fig. \ref{visual_diagram} show some qualitative examples of news image descriptions generated by our method, the variant Non-Rule, and the baseline Tell \cite{TranMX20}. 
To provide insight in how the news-aware semantic rule guides BART model to generate captions, we report the corresponding semantic rule for each example. 
Qualitative results show that the news captions generated by our method can follow the fundamental rules of news reporting, where the individuals and actions associated with the event are accurately captured. We have the following three observations. 

\begin{enumerate}

\item During the construction of semantic rules, we link roles with named entities to guide the caption generation. 
In case (a), 
our approach links the role of “Agent” and the named entity “Lars Bender” by providing explicit affiliations for generic objects and named entities, thereby generating the correct description, ``Lars Bender scored the winning goal for Germany in the 80th minute.” In contrast, due to the lack of rule guidance, both Non-Rule and Tell generate the incorrect named entities “Lukas Podolski”.

\item Our model can capture core events in images to construct the news-aware semantic rule, which is used to accurately depict event-related information to produce captions. For example, in case (c), 
Non-Rule cannot effectively capture the main action ``performing” in the image, while Tell cannot correctly recognize the named entity ``Femi Kuti" associated with the action. So their generated captions are insufficient to accurately describe the image content. 
In contrast, our method effectively extracts the key action ``performing" and its associated entities ``Femi Kuti" using a news-aware semantic rule. We then incorporate this rule into BART, resulting in the generation of accurate news captions. 

\item We notice a failure case in case (d), where our model failed to output information about “Catholic Charities of the Diocese”. The reason is that our semantic rule is designed to capture key events from image, so our model only outputs information ``A fire destroyed the largest halfway house in Paterson, N.J., on Saturday.” In future work, integrating multi-modal knowledge from both images and news articles into rules may have a positive impact on improving the performance of our model. 
\end{enumerate}

\section{Conclusion}
In this paper, we propose a novel rule-driven news captioning method that enables model to follow designated rule signal to generate captions. Specifically, we design the news-aware semantic rule that encompasses the primary action depicted in the image as well as the roles played by the named entities involved in the action. By injecting the rule into the large-scale pre-trained model, BART, using the prefix-tuning strategy, we can generate descriptions that follow the fundamental rules of news reporting. We conduct
experiments on two large-scale public datasets, i.e., GoodNews and NYTimes800k, and the results verify the effectiveness of our method. In future research, we plan to further explore and design the multi-modal rules to generate news descriptions that are more in line with human expectations.

\section*{Acknowledgment}
This work was supported in part by the National Natural Science Foundation of China (U21B2024, 62002257). 

\ifCLASSOPTIONcaptionsoff
  \newpage
\fi

\bibliographystyle{IEEEtran}
\bibliography{IEEEabrv,Bibliography}

\vfill
\end{document}